\documentclass[11pt,a4paper]{article}
\usepackage[hyperref]{acl2020}
\usepackage{times}
\usepackage{latexsym}

\usepackage[normalem]{ulem}
\usepackage{graphicx}
\usepackage{subfigure}
\usepackage{booktabs} 
\usepackage{multirow}
\usepackage{amsmath}
\usepackage{wrapfig}

\usepackage{microtype}

\aclfinalcopy 

\title{\textsc{K-Adapter}: Infusing Knowledge into Pre-Trained Models with Adapters}

\author{
	Ruize Wang$^{1}$\thanks{~~Work is done during internship at Microsoft. }, 
	Duyu Tang$^2$, Nan Duan$^2$, Zhongyu Wei$^1$,
	Xuanjing Huang$^1$, \\ \textbf{Jianshu Ji$^3$, Guihong Cao$^3$, Daxin Jiang$^2$, Ming Zhou$^2$} \\

	$^1$Fudan University, Shanghai, China \\
	$^2$Microsoft, Beijing, China \\
	$^3$Microsoft, Redmond WA, USA \\
    {\tt \{rzwang18,zywei,xjhuang\}@fudan.edu.cn}\\
    {\tt \{dutang,nanduan,jianshuj,gucao,djiang,mingzhou\}@microsoft.com} 
}

\date{}

\begin{document}
\maketitle
\begin{abstract}
We study the problem of injecting knowledge into large pre-trained models like BERT and RoBERTa.
Existing methods typically update the original parameters of pre-trained models when injecting knowledge.
However, when multiple kinds of knowledge are injected, 
the historically injected knowledge would be flushed away.
To address this, we propose \textsc{K-Adapter}, a framework that retains the original parameters of the pre-trained model fixed and supports the development of versatile knowledge-infused model.
Taking RoBERTa as the backbone model, \textsc{K-Adapter} has a neural adapter for each kind of infused knowledge, like a plug-in connected to RoBERTa.
There is no information flow between different adapters, thus multiple adapters can be efficiently trained in a distributed way.
As a case study, we inject two kinds of knowledge in this work, including (1) factual knowledge obtained from automatically aligned text-triplets on Wikipedia and Wikidata and (2) linguistic knowledge obtained via dependency parsing. 
Results on three knowledge-driven tasks, including relation classification, entity typing, and question answering, demonstrate that each adapter improves the performance and the combination of both adapters brings further improvements. 
Further analysis indicates that 
\textsc{K-Adapter} captures versatile knowledge than RoBERTa.
\footnote{Codes are publicly available at \url{https://github.com/microsoft/k-adapter}}
\end{abstract}

\section{Introduction}
Language representation models, which are pre-trained on large-scale text corpus through unsupervised objectives like (masked) language modeling, such as BERT \citep{devlin2018bert}, GPT \citep{radford2018improving,radford2019language}, XLNet \citep{yang2019xlnet}, RoBERTa \citep{liu2019roberta} and T5 \citep{raffel2019exploring}, have established state-of-the-art performances on various NLP downstream tasks. 
Despite the huge success of these pre-trained models in empirical studies, recent studies suggest that models learned in such an unsupervised manner struggle to capture rich knowledge.
For example, \citet{poerner2019bert} suggest that although language models do well in reasoning about the surface form of entity names, they fail in capturing rich factual knowledge.
\citet{kassner2019negated} observe that BERT mostly did not learn the meaning of negation (e.g. ``\textit{not}''). 
These observations motivate us to study the injection of knowledge into pre-trained models like BERT and RoBERTa.


\begin{table*}[ht]
\begin{center}
\resizebox{\linewidth}{!}{%
\begin{tabular}{p{3.8cm}|p{3.5cm}|p{4.7cm}|p{2cm}|p{2.8cm}}
\toprule
\textbf{Model} & \textbf{Knowledge Source }& \textbf{Objective} & \textbf{BERT fixed in training?} & \textbf{Continual knowledge infusion?}\\
\midrule
ERNIE \citep{zhang2019ernie} & Wikipedia, WikiData & entity linking & N & N \\
        \hline
        LIBERT \citep{lauscher2019informing} & WordNet & synonym word prediction, hyponym-hypernym prediction & from scratch & N\\
        \hline
        SenseBERT \citep{levine2019sensebert} & WordNet & word-supersense prediction & from scratch & N \\
\hline
KnowBERT \citep{peters2019knowledge} & Wordnet, Wikipedia, CrossWikis & entity linking , hypernym linking & N & N \\
\hline
WKLM \citep{xiong2019pretrained} & WikiPedia, WikiData & replaced entity detection & N & N \\
\hline
BERT-MK \citep{he2019integrating} & Unified Medical Language System & discriminate between real and fake facts & N & N\\
\hline
K-Adapter~(this work) & Wikipedia, Wikidata, dependency parser & predication prediction, dependency relation prediction & Y & Y \\
\bottomrule
\end{tabular}
}
\vspace{-2mm}

\caption{Comparison between our approach (\textsc{K-Adapter}) and previous works on injecting knowledge into BERT.}
\label{tab:compare-to-other-ki-model}

\end{center}
\vspace{-5mm}
\end{table*}

Recently, some efforts have been made to exploit injecting knowledge into pre-trained language models \citep{zhang2019ernie,lauscher2019informing,levine2019sensebert,peters2019knowledge,he2019integrating,xiong2019pretrained}. Most previous works (as shown in Table \ref{tab:compare-to-other-ki-model}) augment the standard language modeling objective with knowledge-driven objectives and update the entire model parameters.
Although these methods
obtain better performance on downstream tasks, they struggle at supporting the development of versatile models with multiple kinds of knowledge injected \citep{Kirkpatrick2017over}. When new kinds of knowledge are injected, model parameters need to be retrained so that  previously injected knowledge would fade away. 
Meanwhile, the resulting models produce entangled representations, so that it is hard to investigate the effect of each kind of knowledge.


In this paper, we propose \textsc{K-Adapter}, a flexible and simple framework that supports the infusion of multiple kinds of knowledge into large pre-trained models.
\textsc{K-Adapter} 
leaves the original representation of a pre-trained model unchanged and exports different representations for different types of infused knowledge. This is achieved by the integration of compact neural models dubbed adapters. Adapters are knowledge-specific models plugged outside of a pre-trained model, whose inputs are the output hidden-states of intermediate layers of the pre-trained model. 
We take RoBERTa \citep{liu2019roberta} as the base pre-trained model and integrate two types of knowledge, including factual knowledge obtained by aligned Wikipedia text to Wikidata triplets and linguistic knowledge obtained by applying off-the-shell dependency parser to web texts.
In the pre-training phase, we train two adapters independently.
Since adapters have much less trainable parameters compared with RoBERTa, the training process is memory efficient.

We conduct extensive experiments on six benchmark datasets across three knowledge-driven tasks, i.e., relation classification, entity typing, and question answering. Experiments show that \textsc{K-Adapter} consistently performs better than RoBERTa, and achieves state-of-the-art performance on five datasets.
Case study and 
probing experiments 
indicate that \textsc{K-Adapter} captures versatile knowledge than RoBERTa.



\begin{figure*}[!ht]
    \centering
    \includegraphics[width=0.98\textwidth]{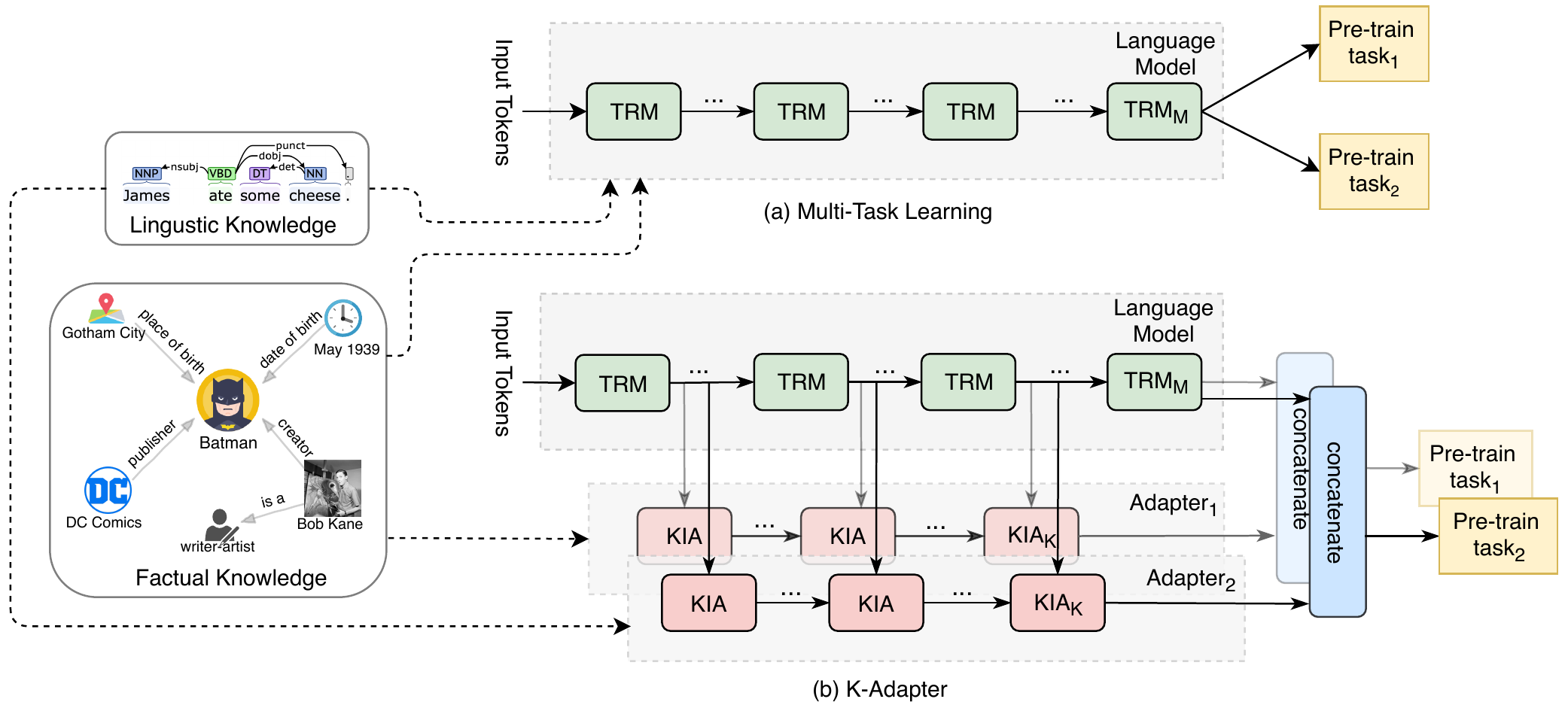}
    \vspace{-2mm}
    \caption{(a) Pre-trained language models inject multiple kinds of knowledge with multi-task learning. Model parameters need to be retrained when injecting new kinds of knowledge, which may result in the catastrophic forgetting (b) Our \textsc{K-Adapter} injects multiple kinds of knowledge by training adapters independently on different pre-train tasks, which supports continual knowledge infusion. When we inject new kinds of knowledge, the existing knowledge-specific adapters will not be affected. KIA represents the adapter layer and TRM represents the transformer layer, both of which are shown in  Figure~\ref{fig:adapter-stru}.}
    \label{fig:overview}
    \vspace{-3mm}
\end{figure*}

\section{Related Work}
\label{related_work}
Our work relates to the area of injecting knowledge into pre-trained models. 
As stated in Table \ref{tab:compare-to-other-ki-model}, previous works mainly differ from the knowledge sources and the objective used for training. 

\textbf{ERNIE} \citep{zhang2019ernie} injects a knowledge graph into BERT. They align entities from Wikipedia sentences to fact triples in WikiData, and discard sentences with less than three entities.
In the training process, the input includes sentences and linked facts, and the knowledge-aware learning objective is to predict the correct token-entity alignment. Entity embeddings are trained on fact triples from WikiData via TransE \citep{bordes2013translating}.
\textbf{LIBERT} \citep{lauscher2019informing} injects pairs of words with synonym and hyponym-hypernym relations in WordNet. The model takes a pair of words separated by a special token as the input, and is optimized by a binary classification problem, which predicts whether the input holds a particular relation or not.
\textbf{SenseBERT} \citep{levine2019sensebert} considers word-supersense knowledge. 
It inject knowledge by predicting the supersense of the masked word in the input, where the candidates are nouns and verbs and the ground truth comes from WordNet.
\textbf{KnowBERT} \citep{peters2019knowledge} incorporates knowledge bases into BERT using Knowledge attention and recontextualization, where the knowledge comes from synset-synset and lemma-lemma relationships in WordNet, and entity linking information in Wikipedia. If entity linking supervision is available, the model is learned with an additional knowledge-aware log-likelihood or max-margin objective.
\textbf{WKLM} \citep{xiong2019pretrained} replaces entity mentions in the original document with names of other entities of the same type.
The model is trained to distinguish the correct entity mention from randomly chosen ones.
\textbf{BERT-MK} \citep{he2019integrating} integrates fact triples from knowledge graph. For each entity, it sample incoming and outcoming instances from the neighbors on the knowledge graph, and replaces head or tail entity to create negative instances. The model is learned to discriminate between real and fake facts.

As shown in Table \ref{tab:compare-to-other-ki-model}, our model (\textsc{K-Adapter}) differs from previous studies in three aspects. First, we consider both fact-related objective (i.e. predicate/relation prediction) and linguistic-related objective (i.e. dependency relation prediction). 
Second, the original parameter of BERT is clamped in the knowledge infusion process. Third, our approach supports continual learning, which means that the learning of different adapters are not entangled. This flexibility enables us to efficiently inject different types of knowledge independently, and inject more types of knowledge without any loss on the previously injected knowledge.

\section{\textsc{K-Adapter}}
As illustrated in Figure \ref{fig:overview} (a), most of the previous works enhance pre-trained language models by injecting knowledge and update model parameters through multi-task learning. Regardless of these different versions of knowledge-injected methods with multi-task learning, common issues not fully studied are catastrophic forgetting of previous knowledge. To address this, we present \textsc{K-Adapter} as shown in Figure \ref{fig:overview}(b), where multiple kinds of knowledge are injected into different compact neural models (i.e., adapters in this paper) individually instead of directly injecting knowledge into pre-trained models. It keeps the original representation of a pre-trained model fixed and supports continual knowledge infusion, i.e., injecting each kind of knowledge into the corresponding knowledge-specific adapter and producing disentangled representation. Specifically, adapters are knowledge-specific models (with few parameters) plugged outside of a pre-trained model. The inputs of adapters are the output hidden-states of intermediate layers of the pre-trained model. Each adapter is pre-trained independently on different tasks for injecting discriminative knowledge while the original parameters of the pre-trained model are frozen. In this paper, we exploit RoBERTa \citep{liu2019roberta} as the pre-trained model, and mainly infuse factual knowledge and linguistic knowledge with two kinds of adapters, i.e., factual adapter and linguistic adapter which are pre-trained on the relation classification task and dependency relation prediction task respectively. In this section, we first describe the structure of our adapter, and then present the process of pre-training knowledge-specific adapters.

\begin{figure}[t!]
    \centering
    \includegraphics[width=0.98\linewidth]{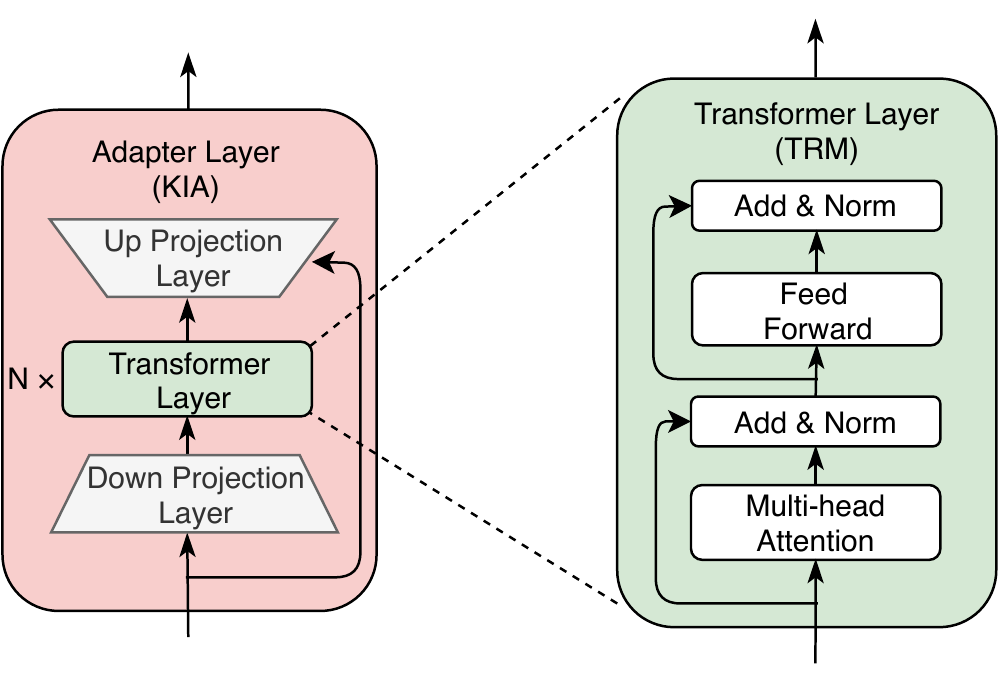}
    \caption{Structure of the adapter layer (left). The adapter layer consists of two projection layers and $N$=2 transformer layers, and a skip-connection between two projection layers.}
    \label{fig:adapter-stru}
    \vspace{-3mm}
\end{figure}

\subsection{Adapter Structure}
In this work, we present a different adapter structure as shown in Figure \ref{fig:adapter-stru}, which is referred to as the knowledge-specific adapter. 
In contrast to \citet{houlsby2019parameter} add adapter layers into each transformer layer, our adapter works as outside plug-ins. Each adapter model consists of $K$ adapter layers that contain $N$ transformer \citep{vaswani2017attention} layers and two projection layers. A skip-connection is applied across two projection layers. Specifically, for each adapter model, we plug adapter layers among different transformer layers of the pre-trained model. We concatenate the output hidden feature of the transformer layer in the pre-trained model and the output feature of the former adapter layer, as the input feature of the current adapter layer. 
For each knowledge-specific adapter, we concatenate the last hidden features of the pre-trained model and adapter as the final output feature of this adapter model. 

In the pre-training procedure, we train each knowledge-specific adapter on different pre-training tasks individually. For various downstream tasks, \textsc{K-Adapter} can adopt the fine-tuning procedure similar to RoBERTa and BERT. When only one knowledge-specific adapter is adopted, we can take the final output feature of this adapter model as the input for task-specific layers of the downstream task. When multiple knowledge-specific adapters are adopted, we concatenate the output features of different adapter models as the input for task-specific layers of the downstream task.

\subsection{Pre-training settings}
We use $\text{RoBERTa}_{LARGE}$ (L=24, H=1024, A=16, 355M params) implementation by Huggingface\footnote{https://github.com/huggingface/transformers} as the pre-trained model in all our experiments. 
As for each adapter layer, we denote the number of transformer layer as $N$, the hidden dimension of transformer layer as $H_A$, the number of self-attention heads as $A_A$, the hidden dimension of down-projection and up-projection layers as $H_d$ and $H_u$.
In detail, we have the following adapter size: $N=2$, $H_A=768$, $A_A=12$, $H_u=1024$ and $H_d=768$. The RoBERTa layers where adapter layers plug in are \{0,11,23\}, and different adapter layers do not share parameters. Thus the total parameters for each adapter model are about 42M, which are much smaller than $\text{RoBERTa}_{LARGE}$ and make the training process memory efficient. It should be noticed that RoBERTa is fixed during training and the parameters of adapters are trainable and initialized randomly. Then we describe how to inject different knowledge into knowledge-specific adapters as below.

\subsection{Factual Adapter}
Factual knowledge can be described as the basic information that is concerned with facts. 
In this work, we acquire factual knowledge from the relationships among entities in natural language.
We extract a sub-dataset T-REx-rc from T-REx \citep{ElSahar18trex} which is a large scale alignment dataset between Wikipedia abstracts and Wikidata triples. We discard all relations having less than 50 entity pairs, collecting 430 relations and 5.5M sentences.
In order to inject factual knowledge, we propose to pre-train a knowledge-specific adapter called facAdapter on the relation classification task. This task requires a model to classify relation labels of given entity pairs based on context. Specifically, the last hidden features of RoBERTa and facAdapter are concatenated as the input representation, and the pooling layer is applied to the input representations of the given entities. Then, we concatenate two entity representations to perform relation classiﬁcation.



\begin{table*}[!t]
\centering
\small
\begin{tabular}{@{}l|ccc|ccc@{}}
\toprule
\multirow{2}{*}{\textbf{Model}} & \multicolumn{3}{c|}{\textbf{OpenEntity}} & \multicolumn{3}{c}{\textbf{FIGER}} \\ \cmidrule(l){2-7} 
 & \textbf{P} & \textbf{R} & \textbf{Mi-$\textbf{F}_1$} & \textbf{Acc} & \textbf{Ma-$\textbf{F}_1$} & \textbf{Mi-$\textbf{F}_1$} \\ \midrule
NFGEC \citep{shimaoka2016attentive} & 68.80 & 53.30 & 60.10 & 55.60 & 75.15 & 71.73 \\ 
BERT-base \citep{zhang2019ernie} & 76.37 & 70.96 & 73.56 & 52.04 & 75.16 & 71.63 \\
ERNIE \citep{zhang2019ernie} & 78.42 & 72.90 & 75.56 & 57.19 & 75.61 & 73.39 \\
KnowBERT \citep{peters2019knowledge} & 78.60 & 73.70 & 76.10 & - & - & - \\
KEPLER \citep{wang2019kepler} & 77.20 & 74.20 & 75.70 & - & - & - \\
WKLM \citep{xiong2019pretrained} & - & - & - & 60.21 & 81.99 & 77.00 \\ \midrule
RoBERTa & 77.55 & 74.95 & 76.23 & 56.31 & 82.43 & 77.83 \\
RoBERTa + multitask & 77.96 & 76.00 & 76.97 & 59.86 & 84.45 & 78.84 \\

\textsc{K-Adapter} (w/o knowledge) & 74.47 & 74.91 & 76.17 & 56.93 & 82.56 & 77.90 \\
\textsc{K-Adapter} (F) & 79.30 & 75.84 & 77.53 & 59.50 & 84.52 & 80.42 \\
\textsc{K-Adapter} (L) & 80.01 & 74.00 & 76.89 & 61.10 & 83.61 & 79.18 \\
\textsc{K-Adapter} (F+L) & 78.99 & 76.27 & \textbf{77.61} & 61.81 & 84.87 & \textbf{80.54} \\
 \bottomrule
\end{tabular}%
\caption{Results on two entity typing datasets OpenEntity and FIGER.}
\label{tab:entity_typing}
\vspace{-2mm}
\end{table*}

\subsection{Linguistic Adapter}
Linguistic knowledge is implicitly contained in natural language texts, e.g., syntactic and semantic information. In this work, we acquire linguistic knowledge from dependency relationships among words in natural language text. 
We build a dataset consisting of 1M examples. 
In particular, we run the off-the-shell dependency parser from Stanford Parser\footnote{http://nlp.stanford.edu/software/lex-parser.html} on a part of Book Corpus \citep{Zhu2015bookcorpus}. 
To inject linguistic knowledge, we pre-train another knowledge-specific adapter called linAdapter on the task of dependency relation prediction. This task aims to predict the head index of each token in the given sentence. 
We concatenate the last hidden features of RoBERTa and linAdapter as the input representation, and then apply a linear layer to input representations of each token to perform classification. 
More training details of facAdapter and linAdapter can be found in the Appendix.


\section{Experiments}
We evaluate our \textsc{K-Adapter} on three knowledge-driven downstream tasks, i.e., entity typing, question answering and relation classification. Furthermore, we conduct detailed analyses with the case study and probing experiments to explore the effectiveness and ability of models for learning factual knowledge. The notations of \textsc{K-Adapter} (F+L), \textsc{K-Adapter} (F), and \textsc{K-Adapter} (L) denote our model which consists of both factual adapter and linguistic adapter, only factual adapter and only linguistic adapter, respectively. 
Implementation details, and statistics of datasets are in the Appendix. 

\subsection{Entity Typing}
We conduct experiments on fine-grained entity typing which aims to predict the types of a given entity and its context. We evaluate our models on OpenEntity \citep{choi2018openentity} and FIGER \citep{ling2015figer} following the same split setting as \citet{zhang2019ernie}. 
To fine-tune our models for entity typing, we modify the input token sequence by adding the special token ``@'' before and after a certain entity, then the first ``@'' special token representation is adopted to perform classification. 
As for OpenEntity, we adopt micro $F_1$ score as the final metric to represent the model performance. As for FIGER, we adopt strict accuracy, loose macro, loose micro $F_1$ scores \citep{lingW12} for evaluation following the same evaluation criteria used in previous works. 

\begin{table*}[!t]
\begin{center}
\begin{small}
\begin{tabular}{@{}l|cc|cc|c@{}}
\toprule
\multirow{2}{*}{\textbf{Model}} & \multicolumn{2}{c|}{\textbf{SearchQA}} & \multicolumn{2}{c|}{\textbf{Quasar-T}} & \textbf{CosmosQA} \\ \cmidrule(l){2-6} 
                                         & \textbf{EM}    & \textbf{$\textbf{F}_1$}    & \textbf{EM}    & \textbf{$\textbf{F}_1$}    & \textbf{Accuracy} \\ \midrule
BiDAF \citep{Seo2016BidirectionalAF}                 & 28.60          & 34.60          & 25.90          & 28.50          & -                 \\
AQA \citep{Buck2017AskTR}                  & 40.50          & 47.40          & -              & -              & -                 \\
R\textasciicircum{}3 \citep{wang2017reinforced} & 49.00          & 55.30          & 35.30          & 41.70          & -                 \\
DSQA \citep{lin2018denoising}                  & 49.00          & 55.30          & 42.30          & 49.30          & -                 \\
Evidence Agg. \citep{wang2017evidence}        & 57.00          & 63.20          & 42.30          & 49.60          & -                 \\
BERT \citep{xiong2019pretrained}                & 57.10          & 61.90          & 40.40          & 46.10          & -                 \\
WKLM \citep{xiong2019pretrained}                & 58.70          & 63.30          & 43.70          & 49.90          & -                 \\
WKLM + Ranking \citep{xiong2019pretrained}       & 61.70          & 66.70          & 45.80          & 52.20          & -                 \\ \midrule
$\text{BERT-FT}_{RACE+SWAG}$ \citep{huang2019cosmos}                                  & -              & -              & -              & -              & 68.70             \\ \midrule
RoBERTa                                  & 59.01          & 65.62          & 40.83          & 48.84          & 80.59             \\
RoBERTa + multitask                      & 59.92              & 66.67              & 44.62              & 51.17              & 81.19             \\

\textsc{K-Adapter} (F)                        & 61.85          & 67.17          & 46.20 & 52.86 & 80.93             \\
\textsc{K-Adapter} (L)                         & 61.15          & 66.82          & 45.66          & 52.39          & 80.76             \\
\textsc{K-Adapter} (F+L)                               & \textbf{61.96} & \textbf{67.31} & \textbf{46.32}          & \textbf{53.00}          & \textbf{81.83}    \\ \bottomrule
\end{tabular}

\caption{Results on question answering datasets including: CosmosQA, SearchQA and Quasar-T.}
\label{tab:open-domain-qa}
\vspace{-2mm}
\end{small}
\end{center}
\end{table*}

\paragraph{Baselines}
\label{exper:entity_typing}
\textbf{NFGEC \citep{shimaoka2016attentive}} employs attentive recursive neural networks to compose context representations. 
\textbf{KEPLER \citep{wang2019kepler}} integrates factual knowledge with the supervision of the knowledge embedding objective. 
\textbf{RoBERTa+multitask} is our RoBERTa model pre-trained with multi-task learning (as shown in Figure \ref{fig:overview}(a)) for injecting multiple kinds of knowledge on two pre-training tasks. 
\textbf{\textsc{K-Adapter} (w/o knowledge)} consists of a RoBERTa model and an adapter without being injected knowledge.
Other baseline models are described in Section \ref{related_work}.

\paragraph{Results and Discussion} 
The results on OpenEntity and FIGER are shown in Table \ref{tab:entity_typing}. 
 \textsc{K-Adapter} (F+L) achieves consistent improvements across these datasets. 
As for OpenEntity, our RoBERTa achieve better results than other baseline models. \textsc{K-Adapter} (F+L) further achieves improvement of 1.38\% $F_1$ over RoBERTa, which means factual knowledge and linguistic knowledge help to predict the types more accurately.
As for FIGER, it covers more entity types, and is more fine-grained than OpenEntity. Compared with WKLM, \textsc{K-Adapter} (F+L) improves the macro $F_1$ by 2.88\%, micro $F_1$ by 2.54\% and accuracy by 1.60\%. This demonstrates that \textsc{K-Adapter} (F+L) benefits fine-grained entity typing.

In addition, we further conduct several experiments on our ablated model \textsc{K-Adapter} (w/o knowledge), to explore whether the performance gains came from introducing knowledge or additional parameters.
Results show that \textsc{K-Adapter} (F) significantly outperforms \textsc{K-Adapter} (w/o knowledge). Moreover, it is worth noting that on OpenEntity dataset, \textsc{K-Adapter} (w/o knowledge) even performs slightly worse than RoBERTa. 
These results demonstrate that our model gains improvement from knowledge instead of more parameters. Thus, for simplicity, we don't discuss \textsc{K-Adapter} (w/o knowledge) in the following experiments.

%


\subsection{Question Answering}
We conduct experiments on two question answering (QA) tasks, i.e., commonsense QA and open-domain QA. Commonsense QA aims to answer questions with commonsense. We adopt CosmosQA \citep{huang2019cosmos} to evaluate our models. CosmosQA requires commonsense-based reading comprehension, formulated as multiple-choice questions. To fine-tune our models for CosmosQA, the input token sequence is modified as \textit{``$<$SEP$>$context $<$/SEP$>$question$<$/SEP$>$answer$<$/SEP$>$''}, then the representation of the first token is adopted to perform classification, and will get a score for this answer. After getting four scores, the answer with the highest score will be selected. We report accuracy scores obtained from the leaderboard. 

Open-domain QA aims to answer questions using external resources such as collections of documents and webpages. We evaluate our modes on two public datasets, i.e., Quasar-T \citep{dhingra2017quasar} and SearchQA \citep{dunn2017searchqa}. 
Specifically, we first retrieve paragraphs corresponding to the question using the information retrieval system and then extract the answer from these retrieved paragraphs through the reading comprehension technique. Following previous work\citep{lin2018denoising}, we use the retrieved paragraphs provided by \citet{wang2017gated} for these two datasets. To fine-tune our models for this task, the input token sequence is modified as \textit{``$<$SEP$>$question $<$/SEP$>$paragraph$<$/SEP$>$''}. We apply linear layers over the last hidden features of our model to predict the start and end position of the answer span. We adopt two metrics including ExactMatch (EM) and loose $F_1$ \citep{lingW12} scores to evaluate our models.

\paragraph{Baselines}
\textbf{$\text{BERT-FT}_{RACE+SWAG}$ \citep{huang2019cosmos}} is the BERT model sequentially fine-tuned on both RACE and SWAG datasets. 
\textbf{BiDAF \citep{Seo2016BidirectionalAF}} adopts a bi-directional attention network.
\textbf{AQA \citep{Buck2017AskTR}} proposes to re-write questions and aggregate the answers generated by the re-written questions.
\textbf{R\textasciicircum{}3 \citep{wang2017reinforced}} is a reinforced model making use of a ranker for selecting most confident paragraph. 
\textbf{Evidence Agg. \citep{wang2017evidence}} proposes making use of the aggregated evidence from across multiple paragraphs. 
\textbf{WKLM \citep{xiong2019pretrained}} is adopted as the reader model to read multiple paragraphs to predict a single answer.
\textbf{WKLM + Ranking \citep{xiong2019pretrained}} is a WKLM paragraph reader plus with a BERT based paragraph ranker to assign each paragraph a relevance score.

\paragraph{Results and Discussion}
The results on CosmosQA are shown in Table \ref{tab:open-domain-qa}. Compared with $\text{BERT-FT}_{RACE+SWAG}$, our RoBERTa significantly achieves 11.89\% improvement of accuracy. 
Compared to RoBERTa, \textsc{K-Adapter} (F+L) further improves the accuracy by 1.24\%, which indicates that \textsc{K-Adapter} can obtain better commonsense inference ability. Moreover, the performance of ablated \textsc{K-Adapter} models, i.e., \textsc{K-Adapter} (F) and \textsc{K-Adapter} (L) are clearly better than RoBERTa, but slightly lose compared with RoBERTa+multitask. It is notable that \textsc{K-Adapter} (F+L) makes obvious improvement comparing with RoBERTa+multitask. This demonstrates that the combination of multiple knowledge-specific adapters could achieve better performance.

The results for open-domain QA are shown in Table \ref{tab:open-domain-qa}. \textsc{K-Adapter} models achieve better results compared to other baselines. This indicates that \textsc{K-Adapter} models can make full use of the infused knowledge and accordingly benefit understanding the retrieved paragraphs to answer the question. Specifically, on SearchQA, \textsc{K-Adapter} (F+L) makes significant improvement of 4.01\% $F_1$ scores, comparing with WKLM where the ranking scores are not used, and even has a slight improvement as compared to WKLM+Ranking. It is worth noting that \textsc{K-Adapter} models do not consider the confidence of each retrieved paragraph, while WKLM+Ranking utilizes ranking scores from a BERT based ranker. On the Quasar-T dataset, \textsc{K-Adapter} (F+L) also outperforms WKLM by 3.1\% $F_1$ score and slightly outperforms WKLM+Ranking.

\begin{table}[!t]
\begin{center}
\resizebox{\linewidth}{!}{%
\begin{tabular}{@{}lccc@{}}
\toprule
\textbf{Model} & \textbf{P} & \textbf{R} & \textbf{$\textbf{F}_1$} \\ \midrule
C-GCN \citep{zhang2018graph} & 69.90 & 63.30 & 66.40 \\
BERT-base \citep{zhang2019ernie} & 67.23 & 64.81 & 66.00 \\
ERNIE \citep{zhang2019ernie} & 69.97 & 66.08 & 67.97 \\
BERT-large \citep{soares2019matching} & - & - & 70.10 \\
BERT+MTB \citep{soares2019matching} & - & - & 71.50 \\
KnowBERT \citep{peters2019knowledge} & 71.60 & 71.40 & 71.50 \\
KEPLER \citep{wang2019kepler} & 70.43 & 73.02 & 71.70 \\ \midrule
RoBERTa & 70.17 & 72.36 & 71.25 \\
RoBERTa +  multitask & 70.18 & 73.11 & 71.62 \\

\textsc{K-Adapter} (F) & 69.39 & 74.59 & 71.89 \\
\textsc{K-Adapter} (L) & 68.85 & 75.37 & 71.96 \\
\textsc{K-Adapter} (F+L) & 70.14 & 74.04 & \textbf{72.04} \\ 
\bottomrule
\end{tabular}%
}
\end{center}
\caption{Results on the relation classification dataset TACRED.}
\label{tab:tacred}
\end{table}

\begin{table*}[]
\huge
\renewcommand\arraystretch{2.7}
\label{tab:case_study_compare}
\resizebox{\textwidth}{!}{%
\begin{tabular}{@{}lllll@{}}
\toprule
\textbf{Input} &
  \textbf{True label} &
  \textbf{Model} &
  \textbf{Predicted label} &
  \textbf{Predicted logits} \\ \midrule
\multirow{2}{4.3in}{His former student \underline{Mark Devlin} of the \uwave{University of Pennsylvania} was co-leader of the other , known as the Microwave Anisotropy Telescope .} &
  \multirow{2}{*}{schools\_attended} &
  K-Adapter &
  {[}'schools\_attended', 'no\_relation','founded'{]} &
  {[}12.6, 9.5, 5.2{]} \\ \cmidrule(l){3-5} 
 &
   &
  RoBERTa &
  {[}'no\_relation', 'founded', ''member\_of''{]} &
  {[}9.1, 6.5, 5.0{]} \\ \midrule
\multirow{2}{4.3in}{\underline{Graham} had been in custody in \uwave{Vancouver} , British Columbia , since June .} &
  \multirow{2}{*}{cities\_of\_residence} &
  K-Adapter &
  {[}'cities\_of\_residence', 'countries\_of\_residence', 'no\_relation'{]} &
  {[}13.5,6.8,6.6{]} \\ \cmidrule(l){3-5} 
 &
   &
  RoBERTa &
  {[}'countries\_of\_residence', 'country\_of\_death', 'alternate\_names'{]} &
  {[}7.1, 7.0, 6.8{]} \\ \midrule
\multirow{2}{4.3in}{\underline{Vladimir Ladyzhenskiy} of Russia died after she suffered a \uwave{shock} in the final of the spa world championship in Heinola , a southern city of Finland , on Saturday .} &
  \multirow{2}{*}{cause\_of\_death} &
  K-Adapter &
  {[}'cause\_of\_death','origin','no\_relation'{]} &
  {[}11.0, 7.6, 7.1{]} \\ \cmidrule(l){3-5} 
 &
   &
  RoBERTa &
  {[}'no\_relation', 'cause\_of\_death', 'origin'{]} &
  {[}6.3, 5.9, 5.5{]} \\ \midrule
\multirow{2}{4.3in}{You can't have a good season unless it starts well, '' said \uwave{Bill Martin}, co-founder of \underline{ShopperTrak}, on Saturday .} &
  \multirow{2}{*}{founded\_by} &
  K-Adapter &
  {[}'founded\_by', 'member\_of', 'employee\_of'{]} &
  {[}10.2, 9.3, 7.3{]} \\ \cmidrule(l){3-5} 
 &
   &
  RoBERTa &
  {[}'no\_relation', 'founded\_by', 'employee\_of'{]} &
  {[}10.0, 8.5, 5.4{]} \\ \bottomrule
\end{tabular}%
}
\caption{A case study for \textsc{K-Adapter} and RoBERTa on relation classification dataset TACRED. \underline{Underlines} and \uwave{wavy lines} highlight the subject entities and object entities respectively. We report the top 3 ranked predictions.}
\label{tab:case_study_compare}
\end{table*}

\begin{table*}[ht]
\centering
\small
\resizebox{\textwidth}{!}{%
\begin{tabular}{p{2in}llp{7cm}}
\toprule
\textbf{Query} &
  \textbf{Answer} &
  \textbf{Model} &
  \textbf{Generation} \\ \midrule
\multirow{2}{1.8in}{The native language of Mammootty is \underline{{[}MASK{]}}.} &
  \multirow{2}{*}{Malayalam} &
  $\text{RoBERTa}_{}$ &
  English, Tamil, Hindi, Sanskrit, Arabic, Chinese
\\ \cmidrule(l){3-4} 
 &
   &
  \textsc{K-Adapter} &
  \textbf{Malayalam}, Tamil, Hindi, Mandarin, English
  \\ \midrule
\multirow{2}{*}{Ravens can \underline{{[}MASK{]}}.} &
  \multirow{2}{*}{fly} &
  $\text{RoBERTa}$ &
  win, play, score, lose, run, drink, \textbf{\textbf{fly}}, roll, wait
  \\ \cmidrule(l){3-4} 
 &
   &
  \textsc{K-Adapter} &
  \textbf{fly}, swim, sing, shoot, kill, go, fish, drink, die
  \\ \midrule
\multirow{2}{*}{Sometimes virus causes \underline{{[}MASK{]}}.} &
  \multirow{2}{*}{infection} &
  $\text{RoBERTa}_{}$ &
  cancer, death, illness, blindness, paralysis
\\ \cmidrule(l){3-4} 
 &
   &
  \textsc{K-Adapter} &
  cancer, illness, death, \textbf{infection}, disease
  \\ \midrule
\multirow{2}{1.8in}{Sunshine Coast, British Columbia is located in \underline{{[}MASK{]}}.} &
  \multirow{2}{*}{Canada} &
  $\text{RoBERTa}_{}$ &
  Florida, California, Texas, Hawaii, Mexico
  \\ \cmidrule(l){3-4} 
 &
   &
  \textsc{K-Adapter} &
  \textbf{Canada}, Vancouver, Victoria, BC, Australia
  \\ \midrule
\multirow{2}{1.5in}{iPod Touch is produced by \underline{{[}MASK{]}}.} &
  \multirow{2}{*}{Apple} &
  $\text{RoBERTa}_{}$ &
  \textbf{Apple}, Samsung, Qualcomm, LG, Microsoft
  \\ \cmidrule(l){3-4} 
 &
   &
  \textsc{K-Adapter} &
  \textbf{Apple}, HTC, Samsung, Motorola, Intel
  \\ \bottomrule
\end{tabular}%
}
\caption{Examples of LAMA generation for $\text{RoBERTa}_{LARGE}$ and \textsc{K-Adapter}. The last column reports the top ranked predicted tokens. Correct predictions are in \textbf{bold}.}
\label{tab:lama-example}
\vspace{-2mm}
\end{table*}

\subsection{Relation Classification}
Relation classification aims to determine the correct relation between two entities in a given sentence.
We adopt a large-scale relation classification dataset TACRED \citep{zhang2017tacred}.
To fine-tune our models for this task, we modify the input token sequence by adding special token ``@'' before and after the first entity, adding ``\#'' before and after the second entity. Then the token representations of the first special token ``@'' and ``\#'' are concatenated to perform relation classification. We adopt micro $F_1$ score as the metric to represent the model performance as previous works.

\paragraph{Baselines}
\textbf{C-GCN \citep{zhang2018graph}} employs graph convolutional networks to model dependency trees. 
\textbf{BERT-large \citep{soares2019matching}} is a baseline BERT-large model. 
\textbf{BERT+MTB \citep{soares2019matching}} is a method of training relation representation without supervision from a knowledge base by matching the blanks.
Other baseline models are described in Section \ref{related_work} and \ref{exper:entity_typing}.

\paragraph{Results and Discussion} Table \ref{tab:tacred} shows the performances of different models on TACRED. The results indicate that \textsc{K-Adapter} models significantly outperform all baselines, which directly demonstrate our models can benefit relation classification. In particular, (1) \textsc{K-Adapter} models outperform RoBERTa, which proves the effectiveness of infusing knowledge into pre-trained model with adapters. (2) \textsc{K-Adapter} models gain more improvement compared with \emph{RoBERTa+multitask}. This directly demonstrates injecting knowledge individually in \textsc{K-Adapter} way would help models make full use of knowledge. 

\subsection{Case Study}
Table \ref{tab:case_study_compare} gives a qualitative comparison example between \textsc{K-Adapter} and RoBERTa on relation classification dataset TACRED. The results show that, in most cases, the wrongly predicted logit value of RoBERTa and the logit value of the true label are actually quite close. 
For example, given ``\textit{\underline{New Fabris} closed down \uwave{June 16}}'', RoBERTa predicts ``\text{no\_relation}'', but the true label ``\textit{city\_of\_birth}'' ranks in second place. If a model could correctly predict the relationship between ``\textit{New Fabris}'' and ``\textit{June 16}'', then it needs to know that ``\textit{New Fabris}'' is a company. Thanks to the factual knowledge in \textsc{K-Adapter}, it can help the model from predicting ``\text{no\_relation}'' to predicting the correct category label.

In addition, we utilize a LAMA (LAnguage Model Analysis) probe \citep{petroni2019language} to examine models' ability to memorize factual knowledge. Specifically, the LAMA probing task is under a zero-shot setting, which requires the language model to answer cloze-style questions about relational facts without fine-tuning, For example, given ``Simon Bowman was born in [MASK]'' as the input, models are asked to predict the correct token which is masked. Table \ref{tab:lama-example} shows several examples for the generation of $\text{RoBERTa}_{LARGE}$ and \textsc{K-Adapter} for LAMA queries. From these examples, we can find that the objects predicted by \textsc{K-Adapter} are more accurate, which demonstrate that \textsc{K-Adapter} captures richer factual knowledge than RoBERTa. More details about this probing experiments can be found in the Appendix \ref{probing_experiments} and \ref{probing_experiments_details}.

\section{Conclusion}
In this paper, we propose a flexible and simple approach, called \textsc{K-Adapter}, to infuse knowledge into large pre-trained models. \textsc{K-Adapter} remains the original parameters of pre-trained models unchanged and supports continual knowledge infusion, i.e., new kinds of injected-knowledge will not affect the parameters learned for old knowledge. Specifically, factual knowledge and linguistic knowledge are infused into RoBERTa with two kinds of adapters, which are pre-trained on the relation classification task and dependency relation prediction task, respectively. Extensive experiments on three knowledge-driven downstream tasks demonstrate that the performance of each adapter achieves a significant improvement individually, and even more together. Detailed analyses further suggest that \textsc{K-Adapter} captures richer factual and commonsense knowledge than RoBERTa, and provide insights on the effectiveness of knowledge infusion.
In future work, we will infuse more types of knowledge, and apply our framework to more pre-trained models.

\bibliography{anthology,acl2020}
\bibliographystyle{acl_natbib}

\newpage
\appendix
\section*{Appendix}
\label{sec:appendix}


\section{Probing Experiments}
\label{probing_experiments}
Although \textsc{K-Adapter} models have shown superior performance on knowledge-driven downstream tasks, it does not directly provide insights into whether our models infuse richer factual knowledge. Thus we utilize a LAMA (LAnguage Model Analysis) probe \citep{petroni2019language} to examine the ability to memorize factual knowledge. Specifically, the LAMA probing task is under a zero-shot setting, which requires the language model to answer cloze-style questions about relational facts without fine-tuning, e.g., ``Simon Bowman was born in [MASK]''. The model needs to predict a distribution over a limited vocabulary to replace [MASK]. We report mean precision at one (P@1) macro-averaged over relations.



\begin{table*}[h]
\centering
\small

\resizebox{\linewidth}{!}{%
\begin{tabular}{@{}l|cccccc@{}}
\toprule
\multirow{2}{*}{Corpus} & \multicolumn{6}{c}{Models}                                                        \\ \cmidrule(l){2-7} 
                        & ELMo & ELMo5.5B & TransformerXL & BERT-large & $\text{RoBERTa}_{LARGE}$ & \textsc{K-Apdater} \\ \midrule
LAMA-Google-RE          & 2.2  & 3.1    & 1.8           & 12.1       & 4.8           & 7.0                  \\
LAMA-UHN-Google-RE      & 2.3  & 2.7    & 1.3           & 6.5        & 2.5           & 3.7                  \\ \midrule
LAMA-T-REx              & 0.2  & 0.3    & 19.5          & 33.9       & 27.1          & 29.1                 \\
LAMA-UHN-T-REx          & 0.2  & 0.2    & 12.6          & 26.2       & 20.1          & 23.0                \\ \bottomrule
\end{tabular}
}
\caption{P@1 on LAMA and LAMA-UHN across Google-RE and T-REx corpora. }
\label{tab:lama}
\end{table*}

\paragraph{Settings}
We consider several language models including: ELMo \citep{peters2018elmo}, ELMo5.5B \citep{peters2018elmo}, Transformer-XL \citep{dai2019transformer}, $\text{BERT}_{LARGE}$ and $\text{RoBERTa}_{LARGE}$. We focus on LAMA-GoogleRE and LAMA-T-REx, which are aimed at factual knowledge. We also conduct probe experiments on LAMA-UHN \citep{poerner2019bert}, a more “factual” subset of LAMA, by filtering out queries that are easy to answer from entity names alone. Different models have different vocabulary sizes. To conduct a more fair comparison experiment, we adopt the intersection of vocabularies and let every language model rank only tokens in this vocabulary following \citet{petroni2019language}. For simplicity, we only compare \textsc{K-Apdater} (F) which is infused with factual knowledge, with other baseline models.
\paragraph{Results and Discussion}
Results are shown in Table \ref{tab:lama}. It is surprising that $\text{BERT}_{LARGE}$ performs better than $\text{RoBERTa}_{LARGE}$. There is one possible reason: BERT uses a character-level BPE \citep{gage1994new} vocabulary, while RoBERTa considers byte-level BPE vocabulary. This finding indicates that, although using bytes makes it possible to learn a subword vocabulary that can encode any text without introducing ``unknown'' tokens, it might indirectly harm the model's ability to learn factual knowledge, e.g., some proper nouns may be divided into bytes. 
Thus in the following experiments, we do not take BERT into account.

\textsc{K-Adapter} outperforms other models (except for BERT) by a huge margin. As for LAMA, compared to $\text{RoBERTa}_{LARGE}$, \textsc{K-Adapter} obtains 2.2\% and 1.2\% P@1 improvement across Google-RE and T-REx, respectively. Moreover, compared to $\text{RoBERTa}_{LARGE}$, \textsc{K-Adapter} still achieves better results on LAMA-UHN. The results demonstrate that \textsc{K-Adapter} captures richer factual and commonsense knowledge than RoBERTa.

\section{Pre-Training Details}
\subsection{Factual Adapter}
The pre-trained model is fixed during training and the parameters of the factual adapter are trainable and initialized randomly. The model is trained with cross-entropy loss. To accelerate the training process, we set the max sequence length as 64 as the average sequence length of T-REx-rc is only 22.8. We train the model for 5 epochs using a batch size of 128. We use AdamW to optimize our models with the initial learning rate of 2e-5. We train the model with 4 16G NVIDIA V100 GPUs.

\begin{figure*}[h]
    \centering
    \includegraphics[width=0.6\textwidth]{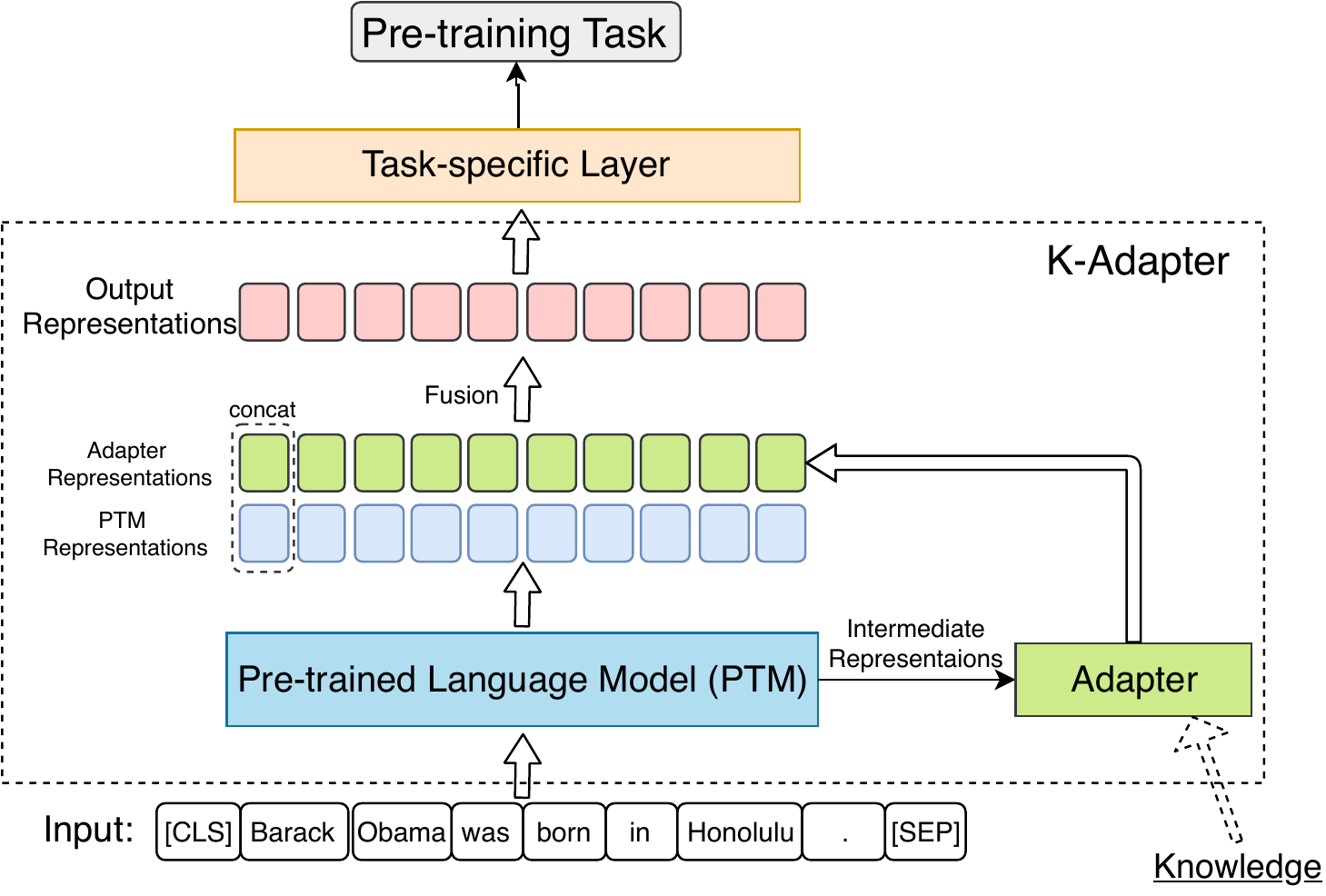}
    \caption{An overview of our \textsc{K-Adapter} to inject specific knowledge by training a knowledge-specific adapter on the pre-training task.}
    \label{fig:kadapter_strucutre_pretraining}
\end{figure*}

\begin{figure*}[h]
    \centering
    \includegraphics[width=0.5\textwidth]{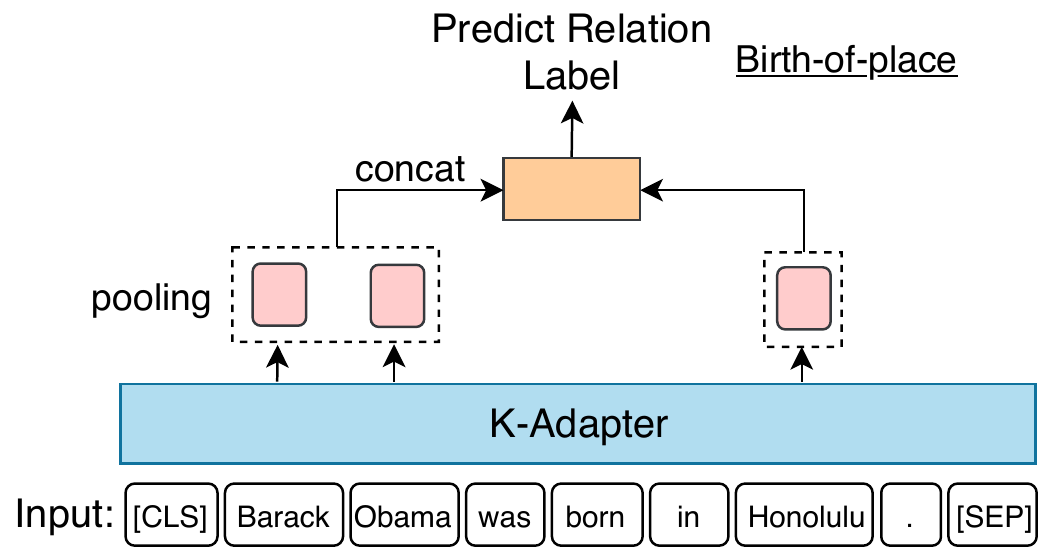}
    \caption{An example of using relation classification as a pre-training task to inject knowledge into \textsc{K-Adapter}: given \textit{ ``Barack Obama was born in Honolulu''}, and then predicts the relationship between ``Barack Obama'' and ``Honolulu'' is ``Birth-of-place''.}
    \label{fig:pretraining_kadapter_relation_clas}
\end{figure*}

\subsection{Linguistic Adapter}
Same as the training process of the factual adapter, the pre-trained model is fixed during training and the parameters of the linguistic adapter are trainable and initialized randomly. The model is trained with BCEWithLogits loss. We set the max sequence length as 128. We train the model for 10 epochs using a batch size of 256. We use AdamW with the initial learning rate of 1e-5. We train the model with 4 16G NVIDIA V100 GPUs.                        
\section{Applying K-adapter on Downstream Tasks}
\begin{figure*}[h]
    \centering
    \includegraphics[width=0.97
    \textwidth]{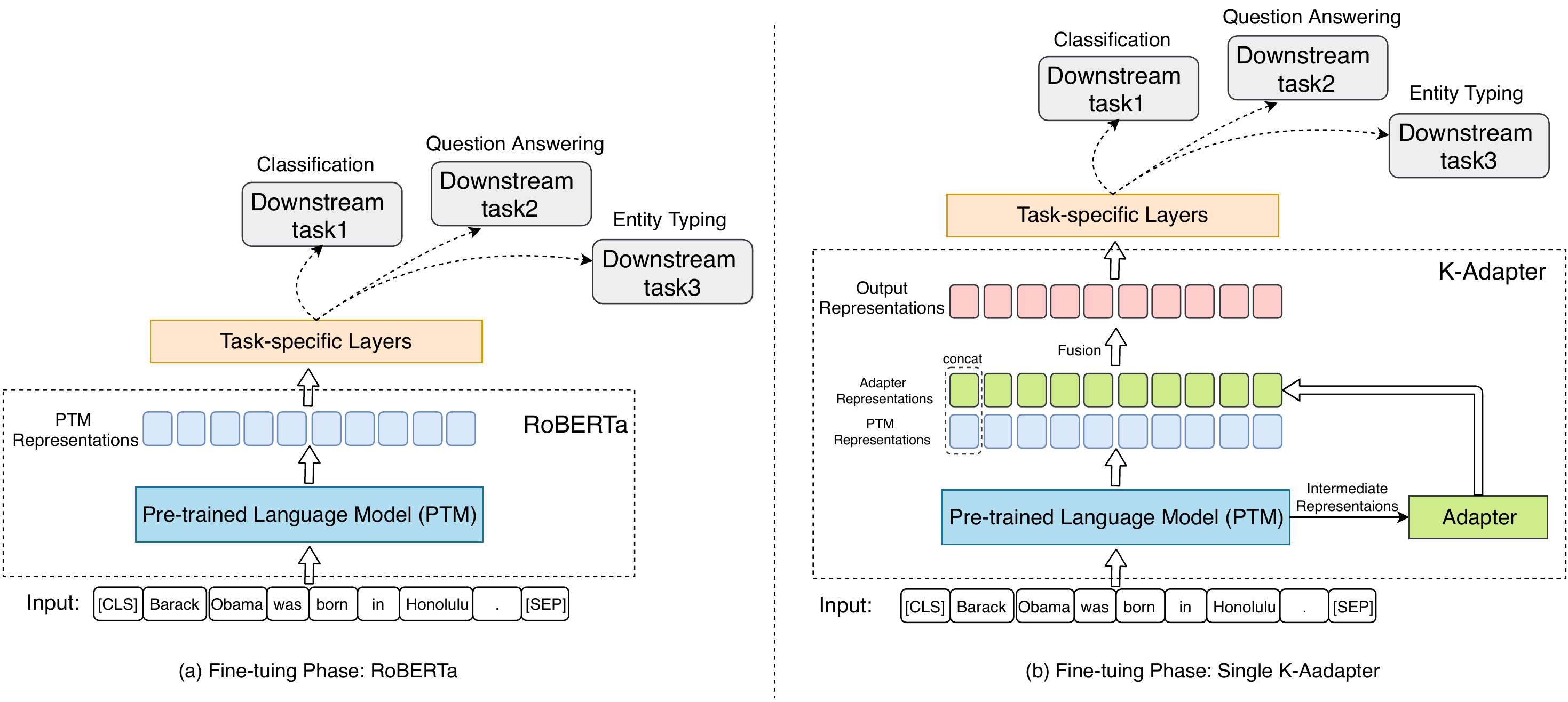}
    \caption{Fine-tuning \textsc{K-Adapter} just like what the original RoBERTa or BERT does.}
    \label{fig:fine_tune_compare}
\end{figure*}

For the downstream tasks, the key point here is the combination of the pre-trained model's representations and adapter's representations, that is to say: leveraging the general information of the pre-trained model on one hand, and the specific knowledge in the adapter on the other.
To use \textsc{K-Adapter} for downstream tasks is very simple as shown in Figure \ref{fig:fine_tune_compare}. Usually, when we use pre-trained language models such as BERT and RoBERTa for downstream tasks, we feed the output features from the pre-trained model into the task-specific layer, and then do the corresponding downstream task.
As for the \textsc{K-Adapter}, we fine-tune it just like what the orginal BERT or RoBERTa does. We concatenate the output features of the pre-trained model with the features of the adapter, and then feed them to the task-specific layer.

\section{Dataset statistics}
In Table \ref{tab:statistics-rc-et}, we present the statistics of one relation classification dataset TACRED, and two entity typing datasets OpenEntity and FIGER. In Table \ref{tab:statistics-qa}, we present the statistics of one commonsense QA dataset CosmosQA and two open-domain QA datasets SearchQA and Quasar-T.

\begin{table}[h]


\begin{center}
\resizebox{\linewidth}{!}{%
\begin{tabular}{@{}l|cccc@{}}
\toprule
Dataset     & Train     & Dev    & Test   & Relation/Type \\ \midrule
TACRED      & 68,124    & 22,631 & 15,509 & 42            \\ \midrule
Open Entity & 2,000     & 2,000  & 2,000  & 6             \\
FIGER       & 2,000,000 & 10,000 & 563    & 113           \\ \bottomrule
\end{tabular}%
}
\end{center}
\caption{The statistics of the relation classification dataset TACRED and entity typing datasets, i.e., Open Entity and FIGER.}
\label{tab:statistics-rc-et}

\end{table}

\begin{table}[h]
\begin{center}
\begin{tabular}{@{}l|ccc@{}}
\toprule
Dataset     & Train     & Dev    & Test \\ \midrule
CosmosQA & 25,588     &3,000  & 7,000   \\ \midrule
SearchQA & 99,811     & 13,893  & 27,247 \\
Quasar-T & 28,496 & 3,000 & 3,000               \\ \bottomrule
\end{tabular}%
\end{center}
\caption{The statistics of the question answering datasets, i.e., CosmosQA, SearchQA and Quasar-T.}
\label{tab:statistics-qa}

\end{table}

\section{Fine-tuning Details}
We implement our experiments using Huggingface\footnote{https://github.com/huggingface/transformers}. For all fine-tuning experiments, we use AdamW as the optimizer. The parameters of adapters are fixed during the fine-tuning process and the parameters of RoBERTa are trainable and initialized from Huggingface checkpoint. We select the best hyperparameters on the validation set. For all experiments, we set the random seed to be 42 for reproductibility.

\subsection{Entity typing}

For Open Entity dataset, we set the max sequence length to be 256 and select the hyperparameters from batch size: \{4, 8\}, learning rate: \{2e-5, 1e-5, 5e-6\} and warmup step: \{0, 200, 500, 1000, 1200\}.
For \textsc{K-Adapter} (F), the best performance is achieved at batch size=4, lr=5e-6, warmup=500 (it takes about 2 hours to get the best result running on singe 16G P100).
For \textsc{K-Adapter} (L), the best performance is achieved at batch size=4, lr=5e-6, warmup=1000 (it takes about 2 hours to get the best result running on singe 16G P100).
For \textsc{K-Adapter} (F+L), the best performance is achieved at batch size=4, lr=5e-6, warmup=1000 (it takes about 3 hours to get the best result running on singe 16G P100). 
For FIGER dataset, we run experiments on 4 16G P100 for 3 epochs, set the max sequence length to be 256, and select the hyperparameters from batch size: \{64, 512, 2048\}, learning rate: \{2e-5, 1e-5, 5e-6\} and warmup step: \{0, 200, 500, 1000, 1200\}.
For \textsc{K-Adapter} (F), the best performance is achieved at batch size=2048, lr=5e-6, warmup=500.
For \textsc{K-Adapter} (L), the best performance is achieved at batch size=2048, lr=5e-6, warmup=200.
For \textsc{K-Adapter} (F+L), the best performance is achieved at batch size=2048, lr=5e-6, warmup=1000.

\subsection{Question Answering}

For CosmosQA dataset, we run experiments on one single 16G P100 for 3 epochs, set the max sequence length to be 256, and select the hyperparameters from batch size: \{16, 32, 64, 128\}, learning rate: \{2e-5, 1e-5, 5e-6\} and warmup step: \{0, 200, 500, 800, 1000\}. 
For \textsc{K-Adapter} (F+L) and its ablated models, the best performance is achieved at batch size=64, lr=1e-5, warmup=0 (it takes about 8 hours to get the best result).

For SearchQA dataset, we run experiments on one single 16G P100 for 2 epochs, set the max sequence length to be 128, and select the hyperparameters from batch size: \{2, 4, 8, 16\}, learning rate: \{5e-5,  2e-5, 1e-5, 5e-6\} and warmup step: \{0, 500, 1000\}. 
For \textsc{K-Adapter} (F+L) and its ablated models, the best performance is achieved at batch size=8, lr=5e-6, warmup=0. For Quasar-T dataset, we run experiments on one single 16G P100 for 5 epochs, set the max sequence length to be 256, and select the hyperparameters from batch size: \{2, 4, 8, 16\}, learning rate: \{5e-5, 2e-5, 1e-5, 5e-6\} and warmup step: \{0, 500, 1000\}. For \textsc{K-Adapter} (F+L) and its ablated models, the best performance is achieved at batch size=16, lr=1e-5, warmup=0.

\subsection{Relation Classification}

For TACRED dataset, we run experiments on 4 16G P100 for 5 epochs, set the max sequence length to be 184, and select the hyperparameters from batch size: \{4, 8, 16, 32\}, learning rate: \{2e-5, 1e-5, 5e-6, 1e-6\} and warmup step: \{0, 200, 500, 800, 1000, 1200\}. 
For \textsc{K-Adapter} (F), the best performance is achieved at batch size=32, lr=1e-5, warmup=500.
For \textsc{K-Adapter} (L), the best performance is achieved at batch size=32, lr=1e-5, warmup=200.
For \textsc{K-Adapter} (F+L), the best performance is achieved at batch size=32, lr=5e-6, warmup=1000.

\subsection{Probing Experiments}
\label{probing_experiments_details}
We implement our probing experiments using LAMA\footnote{https://github.com/facebookresearch/LAMA}.  When we infuse knowledge into knowledge-specific adapters, we do not change the original parameters of the pre-trained model and thus do not adopt the masked language model (MLM) as a pre-training task. Therefore, before we conduct probing experiments, we need to add and train a linear layer as the mlm layer for predicting the [MASK] entities. Specifically, we fix all the parameters of \textsc{K-Adapter} and only update the parameters of the mlm layer using a masked language modeling (MLM) loss. We adopt the raw WikiText-2 dataset (181M). We train the mlm layer with one single 16G P100 for 2 epochs. We set the max sequence length to be 512, batch size to be 1024 and warmup step to be 0.

\end{document}